\documentclass[10pt,twocolumn,letterpaper]{article}

\usepackage{iccv}
\usepackage{times}
\usepackage{epsfig}
\usepackage{graphicx}
\usepackage{amsmath}
\usepackage{amssymb}
\usepackage{multirow}
\usepackage{stfloats}

\usepackage{color, colortbl}
\definecolor{LightCyan}{rgb}{0.88,1,1}

\usepackage[pagebackref=true,breaklinks=true,letterpaper=true,colorlinks,bookmarks=false]{hyperref}

\iccvfinalcopy 

\def\iccvPaperID{8977} 

\ificcvfinal\fi

\begin{document}

\title{\LARGE \bf Best of Both Worlds: Hybrid SNN-ANN Architecture for Event-based Optical Flow Estimation}


\author{Shubham Negi, Deepika Sharma, Adarsh Kumar Kosta and Kaushik Roy\\
Elmore Family School of Electrical and Computer Engineering, Purdue University\\
West Lafayette, IN 47907, USA\\
{\tt\small snegi@purdue.edu}
}

\maketitle
\ificcvfinal\thispagestyle{empty}\fi

\begin{abstract}
In the field of robotics, event-based cameras are emerging as a promising low-power alternative to traditional frame-based cameras for capturing high-speed motion and high dynamic range scenes. This is due to their sparse and asynchronous event outputs. Spiking Neural Networks (SNNs) with their asynchronous event-driven compute, show great potential for extracting the spatio-temporal features from these event streams. In contrast, the standard Analog Neural Networks (ANNs\footnote{We refer to traditional deep learning networks that are non-recurrent in nature as ANNs due to their inputs/computations involving higher bit-precision compared to binary precision in SNNs. The underlying computing hardware is still digital.\label{fn1}}) fail to process event data effectively. However, training SNNs is difficult due to additional trainable parameters (thresholds and leaks), vanishing spikes at deeper layers, and a non-differentiable binary activation function. Furthermore, an additional data structure, ``membrane potential", responsible for keeping track of temporal information, must be fetched and updated at every timestep in SNNs.
To overcome these challenges, we propose a novel SNN-ANN hybrid architecture that combines the strengths of both. Specifically, we leverage the asynchronous compute capabilities of SNN layers to effectively extract the input temporal information. Concurrently, the ANN layers facilitate training and efficient hardware deployment on traditional machine learning hardware such as GPUs. We provide extensive experimental analysis for assigning each layer to be spiking or analog, leading to a network configuration optimized for performance and ease of training. We evaluate our hybrid architecture for optical flow estimation on DSEC-flow and Multi-Vehicle Stereo Event-Camera (MVSEC) datasets. On the DSEC-flow dataset, the hybrid SNN-ANN architecture achieves a 40\% reduction in average endpoint error (AEE) with 22\% lower energy consumption compared to Full-SNN, and 48\% lower AEE compared to Full-ANN, while maintaining comparable energy usage.
\end{abstract}

\begin{figure}[t]
\centering
\includegraphics[width=0.4\textwidth]{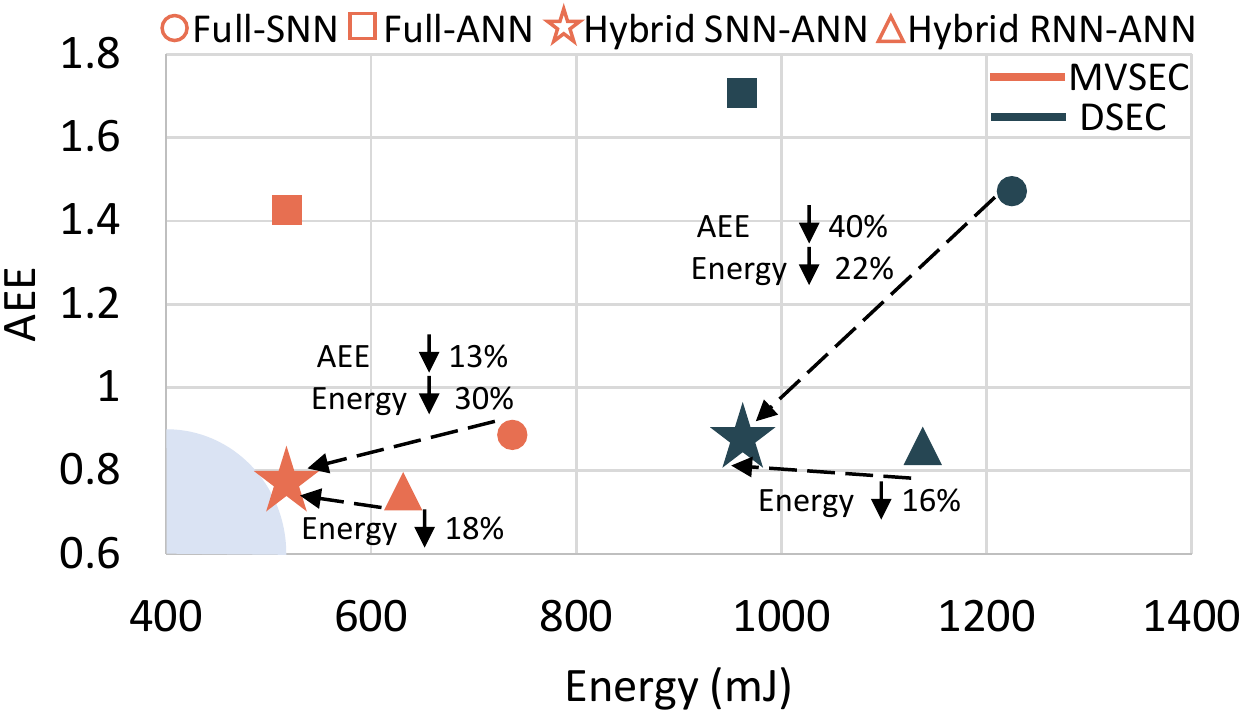}
\caption{AEE (lower is better) vs Inference energy for different versions of EV-FlowNet \cite{evflow} on DSEC-flow and MVSEC dataset deployed on Eyeriss hardware \cite{chen2016eyeriss}. The proposed hybrid SNN-ANN shows lower AEE and energy compared to Full-ANN, SNN and RNN\footref{fn2} counterpart. The shaded blue region shows the preferred region.}
\centering
\label{comparison}
\vskip -0.2 in
\end{figure}

\section{INTRODUCTION}

Robotic systems operating at the edge require visual perception capabilities to perform various tasks, including optical flow estimation and high speed object tracking. These capabilities must be characterized by real-time processing, low energy, and robust sensing \cite{baxi2022towards}. Traditional vision systems face challenges in meeting these demands, especially with high-speed motion and varying lighting conditions. Event-based cameras, such as Dynamic Vision Sensor (DVS) \cite{dvs128, dvs240}, offer a promising alternative. They provide an asynchronous event stream at a high temporal resolution in the order of microseconds, high dynamic range, and extremely low power consumption \cite{posch2014retinomorphic}. However, using Analog Neural Networks (ANNs)\footref{fn1} to directly process the event stream is inefficient due to the stateless computations in ANNs \cite{hagenaars2021self}. Event streams are handled in ANNs either by encoding timing information as input channels~\cite{zhu2019unsupervised}  or by specially curated input encoding techniques~\cite{evflow}. In contrast, machine learning architectures such as Recurrent Neural Networks (RNNs) or Long short-term memory (LSTM) do not require any input encoding and are suitable for sequential tasks \cite{hagenaars2021self}. But these models are more complex with larger model footprints, difficult to train, and have high computation cost ~\cite{ponghiran2021hybrid}.

Spiking Neural Networks (SNNs), which draw inspiration from biological neurons, have emerged as a promising candidate for processing sequential tasks with asynchronous and sparse event stream~\cite{rathi2022exploring}. At the heart of an SNN is the Leaky Integrate and Fire (LIF) neuron, which can be characterized by an internal state, known as the membrane potential \cite{roy2019towards}. The membrane potential in an LIF neuron accumulates the inputs over time and decays at a certain rate governed by the leak at each timestep, emitting a spike whenever the membrane potential exceeds a specified threshold. The accumulation of inputs over time allows SNNs to learn the timing information from the input without any explicit temporal encoding, making them a special case of RNNs with less complexity.

Nonetheless, training deep SNNs remains a challenge ~\cite{rathi2021exploring} due to vanishing spikes in deeper layers, non-differentiable activations and additional threshold and leak parameters. In addition, SNN training is performed using backpropagation through time (BPTT) \cite{werbos1990backpropagation} by unrolling the network in time and propagating errors, layer by layer. This unrolling over sparse activations in SNNs exacerbates vanishing gradients leading to performance degradation and slow convergence. Several solutions have been proposed over the past years to address these challenges, including learnable thresholds/leaks \cite{hagenaars2021self, kosta2022adaptive}, approximate surrogate gradients, \cite{lee2020enabling, rathi2021diet} etc. However, these solutions are still subpar compared to ANN training methodologies.

In addition to the challenges related to training, the efficiency benefits of SNNs can be best obtained when deployed (for inference) on specialized neuromorphic hardware such as IBM's TrueNorth \cite{debole2019truenorth}, Intel's Loihi \cite{davies2018loihi} etc. However, such experimental hardware architectures are yet to scale to large SNN models and are also inaccessible to most researchers. Deploying SNNs on general-purpose hardware such as GPUs, or specialized machine learning accelerators such as Eyeriss \cite{chen2016eyeriss} turns out to be highly inefficient. This is attributed to SNNs requiring an additional data structure for the membrane potential that needs to be fetched and updated at each timestep. This results in excessive data movement and thereby, high inference energy.

To that effect, we propose a novel hybrid SNN-ANN architecture with initial SNN layers at the input, followed by several ANN layers to obtain the best of both worlds. The SNN layers with their inherent recurrence allow capturing temporal information from the sparse and asynchronous event stream. On the other hand, the ANN layers facilitate training and efficient deployment on traditional machine learning hardware. Fig.~\ref{comparison} depicts a comparison between Full-ANN, Full-SNN, Hybrid RNN-ANN\footnote{First layer is ConvRNN (Fig.~\ref{architecture}(c)) and other layers are ANN.\label{fn2}} and the proposed hybrid architecture for the task of optical flow estimation. Our contributions can be summarized as follows:

\begin{itemize}

    \item We introduce a novel hybrid SNN-ANN architecture to efficiently process the sparse spatio-temporal data from event-based cameras (Section~\ref{proposed_arch}).
    \item We analyzed the hybrid SNN-ANN architecture (with $m$ spiking layers), to select a network configuration optimized for both the number ($m$) and position of spiking layers within the network, having the best performance and training complexity for sequential tasks such as optical flow estimation (Section~\ref{intuition}, \ref{dsec}).  
    
    \item We show that our configured hybrid architectures consistently offer lower AEE and energy for the task of optical flow estimation compared to Full-ANN, Full-SNN as well as past hybrid architectures on DSEC-flow \cite{gehrig2021dsec} and MVSEC datasets \cite{zhu2018multivehicle} (Section~\ref{dsec},~\ref{mvsec}).

    \item We provide an estimate of the inference energy using an analytical model (Timeloop~\cite{parashar2019timeloop}) of the underlying hardware (Eyeriss~\cite{chen2016eyeriss}) that includes the cost of both data movement and computations (Section~\ref{exp}).
\end{itemize}

\section{Related Work} \label{relatedwork}
In recent years, there has been an increasing interest in exploiting the high temporal resolution of event cameras to estimate the optical flow using neural network based approaches \cite{meister2018unflow, evflow, ye2020_unsupervised, lee2020spike, lee2022fusion}. For instance, researchers in \cite{evflow} used an ANN-based U-net \cite{unet} for optical flow estimation using self-supervised loss based on grayscale images. Recently, authors in \cite{zhu2019unsupervised} have dealt with time domain information by passing it as a channel to the first layer. With respect to lightweight architectures, authors in \cite{fireflownet} proposed Fire-FlowNet for flow estimation. However, all these approaches require preprocessing to encode timing information in the input. Moreover, passing time domain information as input channels aggravates the computation and parameter overheads.   

In regards to SNN-based architectures, authors in \cite{lee2020spike} proposed Spike-FlowNet, a hybrid version of EV-FlowNet \cite{evflow}, with an SNN encoder and ANN decoder. Further, authors in \cite{lee2022fusion} expanded the framework in \cite{lee2020spike} by adding sensor fusion and a secondary ANN-based encoder for frame-based data. However, the reasoning behind using a full-SNN encoder was lacking in these works. 
To that end, in this work, we first perform an ablation study to understand the impact of the number and position of spiking layers in the hybrid architecture. We demonstrate that a well-designed hybrid SNN-ANN architecture, with the first layer as spiking and subsequent layers as ANNs, achieves superior AEE compared to previous hybrid architecture \cite{lee2020spike} on the task of optical flow estimation. Authors in \cite{hagenaars2021self, kosta2022adaptive} explored trainable threshold and leak using a surrogate gradient for flow estimation. In contrast to these works, we propose a hybrid SNN-ANN architecture that utilizes the benefits of both SNNs and ANNs -- SNNs help to learn temporal information, while ANNs facilitate training and enable efficient mapping to the non-spiking hardware. Further, we show that the hybrid architecture can achieve lower AEE compared to the fully spiking architecture, spiking architectures with explicit recurrence \cite{hagenaars2021self}, and the hybrid architectures proposed in \cite{lee2020spike}. We do not compare our results with Fusion-FlowNet \cite{lee2022fusion} since it uses the inputs from both the event and frame-based cameras to predict the optical flow.

\section{Method} \label{method}

\subsection{Sensor and Input Representation}

\begin{figure}[t]
\centering
\includegraphics[width=0.32\textwidth]{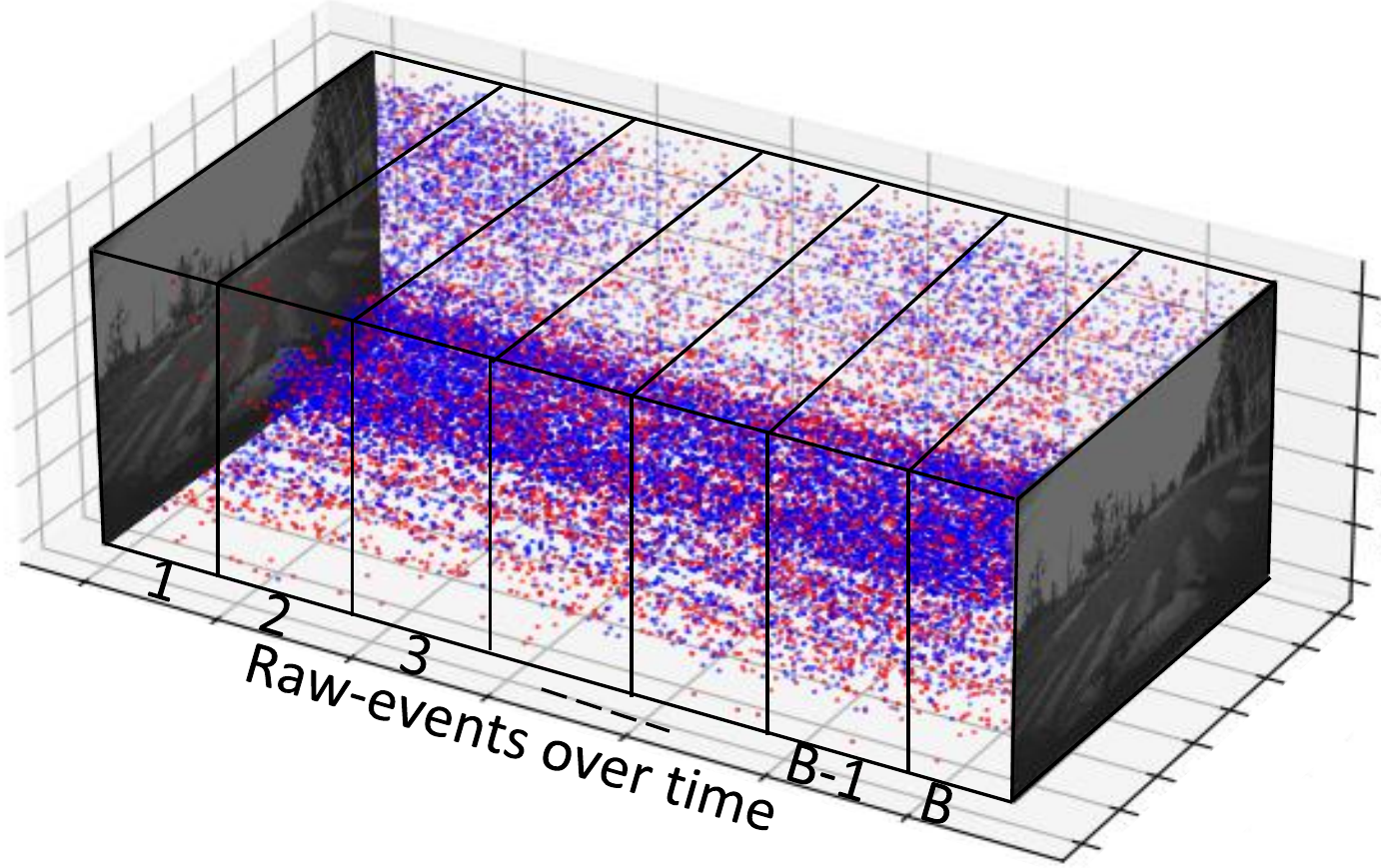}
\caption{Event stream between two grayscale images. The event streams are binned and fed to the networks directly.}
\centering
\label{input_rep}
\vskip -0.2 in
\end{figure}

An event camera responds to intensity changes $(I)$ in the scene asynchronously and independently for every element in the pixel array \cite{gallego2020event}. It generates a discrete event whenever the change in log intensity exceeds a threshold ($\theta$) i.e. $||\log(I_{t+1}) - \log(I_t)||$ $\geq$ $\theta$. The discrete events are generated in the Address Event Representation (AER) format consisting of a tuple $\{x,y,t,p\}$, where $x,y$ are the pixel coordinates; $t$ is the timestamp; and $p$ is the polarity of change. 
Stateless ANNs (i.e. non-recurrent) require encoding temporal information in the input \cite{evflow,fireflownet}. However, in this work, we directly pass inputs to all networks without encoding. For a set of N input events $\{(x_i,y_i,t_i,p_i)\}_{i\in[1,N]}$ between two consecutive grayscale images, a set of B event bins are created using bilinear sampling as shown in Fig.~\ref{input_rep}.


\subsection{Neuron Model}

\begin{figure}[t]
\centering
\includegraphics[width=0.35\textwidth]{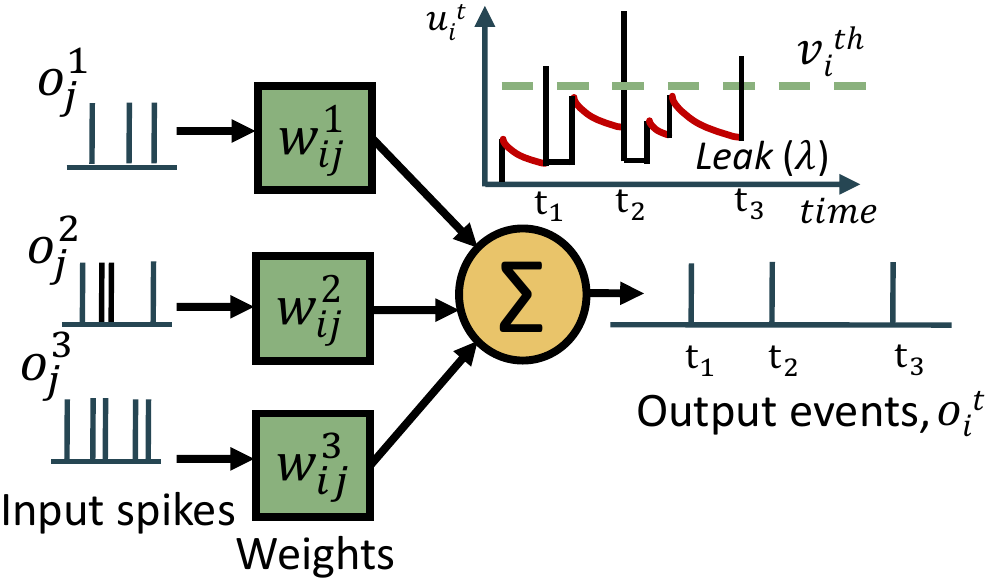}
\caption{Neuronal Dynamics of a LIF neuron.}
\centering
\label{lif}
\vskip -0.2 in
\end{figure}

Spiking neurons are fundamental units in SNNs. In this work, we use the Leaky-Integrate-and-Fire (LIF) \cite{abbott1999lapicque} neuron which accumulates the input information over time into the neuronal state called 'membrane potential' ($u$). Once the membrane potential crosses a specified firing threshold($v_i^{th}$), the neuron fires, generating a spike. Further, the LIF neuron also has an internal mechanism to control forgetting over time using a parameter called leak ($\lambda$). The discretized version of LIF neuron can be described as: 
\begin{equation}\label{eq1}
    u_{i}^t = \lambda u_i^{t-1} + \sum\limits_jw_{ij}o_j^t - v_i^{th}o_i^{t-1}
\end{equation}
\begin{equation}\label{eq2}
    z_i^t = u_i^t/v_i^{th} -1 ,   
    o_i^t = \begin{cases}
    1, & \text{if } z_i^t > 0 \\
    0, & \text{otherwise}
    \end{cases}
\end{equation}

where $u$ is the membrane potential, subscripts $i$ and $j$ represent the post and pre-neurons, respectively, $t$ denotes timesteps, $\lambda$ is the leak factor, $w_{ij}$ is the weight between the i-th and j-th neurons, $o$ is the output spike, and $v_i^{th}$ is the threshold. The first term in Eq.~(\ref{eq1}) denotes the leakage in membrane potential, the second term computes the weighted summation of input spikes, and the third term denotes the reduction of membrane potential upon generation of an output spike at the current layer. This reduction by an amount equal to the firing threshold is termed as a 'soft reset', while a reset to zero is termed as a 'hard reset'. In this work, we use hard reset. Eq.~(\ref{eq2}) shows the spike generation mechanism in LIF neurons as depicted in Fig.~\ref{lif}.

\subsection{Proposed Architecture}
\label{proposed_arch}

The proposed hybrid SNN-ANN architecture is a class of neural network architectures that incorporate both LIF and ReLU neurons in different layers, providing the advantages of both SNN and ANN architectures. The use of hybrid architecture is attributed to two main factors. Firstly, SNNs can directly process the outputs of event cameras, capturing the timing information. Secondly, ANN layers facilitate training and address the challenge of vanishing gradients in SNNs. Lastly, this architecture enables deployment on non-spiking hardware such as GPUs.

\begin{figure*}[t]
\centering
\includegraphics[width=0.75\textwidth]{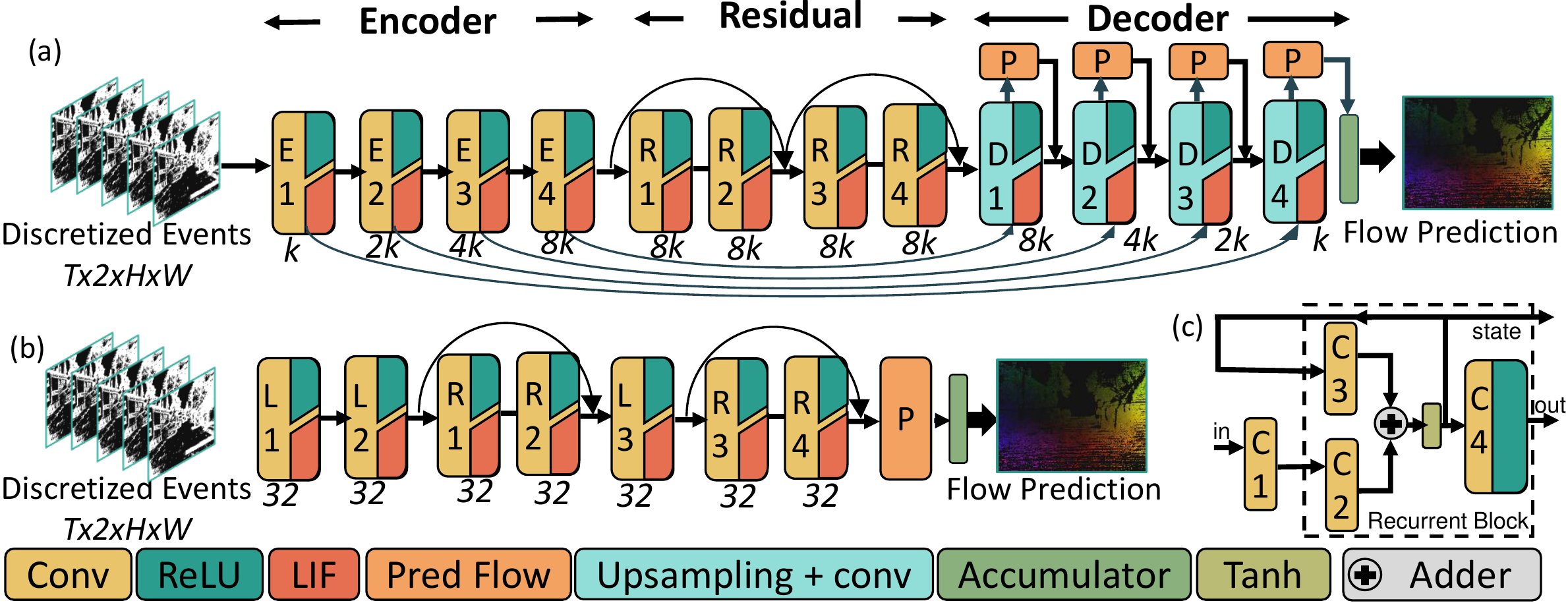}
\caption{Hybrid SNN-ANN architecture based on (a) EV-FlowNet \cite{evflow} and (b) Fire-FlowNet \cite{fireflownet}. The activation layers can be ReLU or LIF, depending on whether the neural network is an ANN, SNN or Hybrid SNN-ANN. Numbers below the layers show the output channels in the layer. In the input dimension, $T$ is the timestep and $2$ is the input channel because of positive and negative polarity. (c) Conv-RNN based recurrent layer \cite{hagenaars2021self}}
\centering
\label{architecture}
\vskip -0.2 in
\end{figure*}

The hybrid architecture can be derived from any non-recurrent ANN for the task of optical flow estimation. We explore two types of hybrid architectures: an encoder-decoder based multi-scale architecture based on EV-FlowNet \cite{evflow} and a lightweight architecture called Fire-FlowNet \cite{fireflownet}. We evaluate several architecture sizes for the first architecture by subsequently scaling down the number of channels by a factor of $2$ at each layer. Both these architectures are shown in Fig.~\ref{architecture}. A forward pass through these architectures involves passing a 2-channeled input sequentially over $T$ timesteps. In the encoder layer (Fig.~\ref{architecture} (a)) the input activation is first downsampled before passing to the next layer. Next, the output from the final encoder layer is passed through two residual blocks and four decoder layers. The decoder layers upsample the input using transposed convolution and also produce intermediate multi-scale flow predictions for each timestep. The flow prediction layers accumulate the flow over all the timesteps to generate the final full-scale flow prediction. Fire-FlowNet architecture (Fig.~\ref{architecture} (b)) performs similar forward passes without any downsampling/upsampling. We also compare with the corresponding RNN\footref{fn2} architecture which uses the ConvRNN layer shown in Fig.~\ref{architecture} (c). The ConvRNN layer requires three extra convolutions, element-wise addition and tanh operation compared to the spiking layer shown in Fig.~\ref{lif}.

\subsection{Optimizing Spiking Layers in Hybrid Architectures} 
\label{intuition}

Upon introducing the hybrid architecture, two questions need to be answered: \noindent (1) How many layers should be spiking? \noindent (2) What should be the positions of the spiking layers? 

To address the first question, we explore the training complexity of hybrid architecture as we increase the number of spiking layers. The accuracy of an SNN is highly sensitive to the additional parameters associated with the spiking neuron (threshold and leak) \cite{rathi2022exploring, hagenaars2021self}. Consequently, increasing the number of spiking layers increases the training complexity making it difficult for the model to converge. 

To address the second question, we consider the characteristics of the event stream and LIF neurons. Data from sensors such as DVS \cite{dvs240} can contain a significant amount of noise from small intensity changes in the static regions of the scene, resulting in events. This can originate from a variety of environmental factors such as reflections, edges due to shadows etc. The leak in LIF neurons serves as a low-pass filter resulting in more robust outcomes for noisy spike inputs \cite{chowdhury2021towards}. If the deeper layers are spiking while the initial layers are non-recurrent ANNs, the noise present in inputs might not be adequately filtered out by the initial ANN layers. This could potentially lead to decreased accuracy. 

Based, on this discussion we posit that a hybrid architecture with LIF neurons occupying few initial layers and all later layers being ReLU would be most effective. We will further validate these intuitions with proper experiments in section ~\ref{dsec}.

\subsection{Loss Functions}
\label{loss}
To show the efficacy of hybrid architecture in learning temporal information from the input, we evaluate it on Multi-Vehicle Stereo Event Camera (MVSEC) \cite{zhu2018multivehicle} and DSEC-flow \cite{gehrig2021dsec} datasets. We use a self-supervised loss using grayscale frames on the MVSEC dataset as it lacks reliable ground truth labels. For DSEC-flow, we utilize a supervised loss. These are described below:

\textbf{Self-supervised Loss}: The self-supervised loss consists of a photometric reconstruction loss ($\mathcal{L}_{photo}$) and a smoothness loss ($\mathcal{L}_{smooth}$) \cite{yu2016back}. Photometric loss is computed using two consecutive grayscale images ($I_{t}(x,y), I_{t+dt}(x,y))$ and the predicted optical flow $(u_{x,y},v_{x,y})$. It is based on the assumption that the brightness is consistent i.e. moving from pixel location $(x,y)$ at time $t$ to pixel location $(x+dx, y+dy)$ at time $(t+dt)$, the intensity remains the same.
The predicted optical flow from the network is used to warp the second 
grayscale image to the first grayscale image \cite{jaderberg2015spatial} and obtain the inverse warped image $I_{t+dt}(x + u, y + v)$.
\begin{equation}\label{eq3}
    \mathcal{L}_{photo} = \sum\limits_{x,y}\rho(I_t(x,y) - I_{t+dt}(x + u, y + v))
\end{equation}
where $\rho$ is Charbonnier loss ($\rho(x) = (x^2 + \eta^2)^r$) used for outlier rejection \cite{sun2014quantitative}. We set r=0.45 and $\eta$=0.001 \cite{lee2020spike}.

The smoothness loss minimizes the difference in optical flow between neighboring pixels and acts as a regularizer.
\begin{equation}\label{eq4}
    \noindent \mathcal{L}_{smooth} = 
    \sum\limits_{x,y} \sum\limits_{i,j\in\mathcal{N}(x,y)}(|u_{x,y}-u_{i,j}|+|v_{x,y}-v_{i,j}|)\\   
\end{equation}
where $\mathcal{N}(x,y)$ is the neighborhood of $(x,y)$. The overall loss is:

\begin{equation}\label{eq5}
    \mathcal{L}_{total} = \mathcal{L}_{photo} + \lambda\mathcal{L}_{smooth}
\end{equation}

where $\lambda$ is the weight factor. 


\textbf{Supervised Loss}: The supervised mean squared error (MSE) loss between the predicted flow $(u_{pred}, v_{pred})$ and the ground truth flow $(u_{gt}, v_{gt})$ is computed as:
\begin{equation} \label{eq6}
    \mathcal{L}_{supervised} = \frac{1}{K}\sum\limits_{i=1}^K((u_{pred} - u_{gt})^2 + (v_{pred} - v_{gt})^2)   
\end{equation}
where K is the number of pixels with non-zero flow.

\subsection{Training Hybrid Architecture}

The training process for SNNs and hybrid architecture differs from that of ANNs due to the non-differentiable hard threshold used as the activation function (LIF). To overcome this challenge, the training methodology proposed in \cite{kosta2022adaptive, lee2020enabling} is utilized, which involves using a surrogate gradient approximation to compute the gradients of the LIF neuron. Additionally, the network is unrolled in time for all timesteps, and the loss is backpropagated using the surrogate gradient and Backpropagation Through Time (BPTT) \cite{werbos1990backpropagation}. Apart from training the weights for SNNs and hybrid architectures, the threshold and leak parameters are also trained to enhance convergence for these networks. To get faster convergence for all the networks we also use batch normalization through time (BNTT) \cite{bntt} during training.

\section{Experiments and Results}
\label{exp}

\subsection{Experimental Setup}
\label{exp_setup}

\begin{figure*}[t]
\centering
\includegraphics[width=0.95\textwidth]{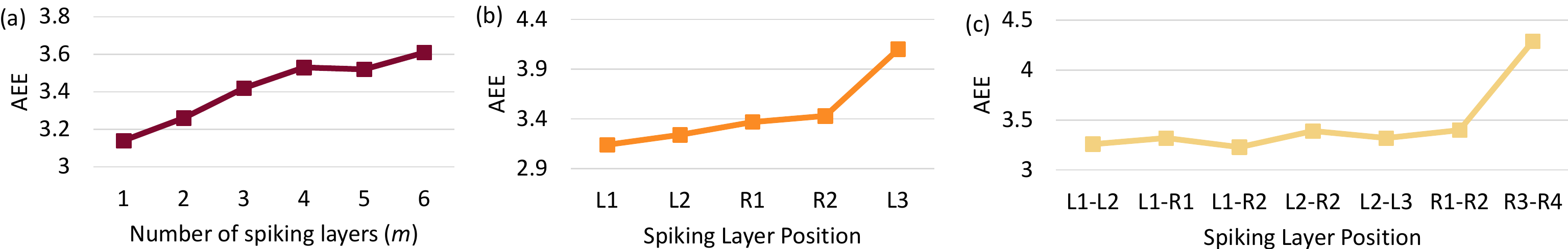}
\caption{Ablation study results on DSEC-flow dataset for (a) number of spiking layers (b) spiking layer position (c) spiking layer position with 2 spiking layers in the model. L1, L2, R1, R2, L3 are the layers in Fire-FlowNet architecture from Fig.~\ref{architecture}(b).}
\centering
\label{ablation}
\vskip -0.2 in
\end{figure*}

\textbf{DSEC-Flow}: The DSEC-Flow dataset \cite{gehrig2021dsec} consists of optical flow ground truths for 24 driving sequences for a total of 7800 training samples and 2100 test samples. However, the ground truths for the test sequences are not public. Therefore we create our own test sequence by splitting the train set in an 80:20\% split for the train and test set \cite{ponghiran2022event}. We train the networks for 200 epochs using adam optimizer \cite{kingma2014adam} with a multi-step learning rate scheduler. We start with a learning rate of 0.001 and scale it by 0.7 every 10 epochs. The training is performed on a single NVIDIA A40 GPU in a supervised fashion using the loss function from Section.~\ref{loss}.

\textbf{MVSEC}: The MVSEC dataset \cite{zhu2018multivehicle} comprises of four indoor flying sequences, two outdoor day driving sequences, and three outdoor night driving sequences. The networks are trained in a self-supervised learning fashion. Training is performed on the outdoor\_day2 driving sequence and evaluation is performed on the indoor\_flying1,2,3 sequences, as well as the outdoor\_day1 sequence. 

Transformations involving random flipping, rotation and cropping to 256x256 size are applied. The learning rate scheduler and optimizer are similar to DSEC experiments. We train for 100 epochs with an initial learning rate of 0.01. The training and evaluation are done on event volumes associated with consecutive pairs of grayscale images (termed as dt=1).

\textbf{Optical Flow Evaluation Metric}: 
To evaluate optical flow, for both datasets, we use average endpoint error (AEE). AEE measures the mean distance between the predicted and ground truth flow. This is done only for pixels containing events ($n$). AEE is calculated as shown below: 

\begin{equation}\label{aee}
    AEE = \frac{1}{n}\sum\limits_{n}|| (u,v)_{pred} - (u,v)_{gt} ||_{2}
\end{equation}

\begin{figure*}[t]
\centering
\includegraphics[width=0.9\textwidth]{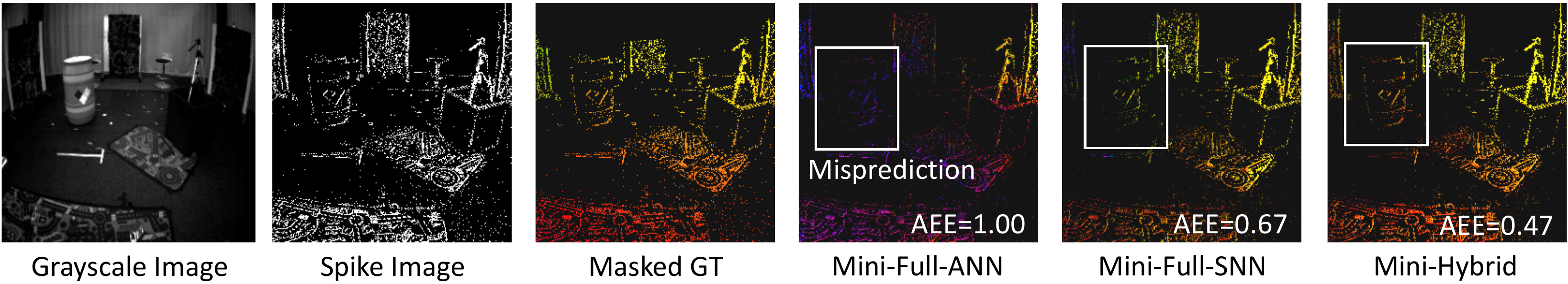}
\caption{Qualitative results of best performing Full-ANN, SNN and Hybrid SNN-ANN on indoor\_flying1 sequence from MVSEC dataset.}
\centering
\label{qualitative}
\end{figure*}

\textbf{Energy Estimation}: Several recent works for optical flow estimation \cite{lee2020spike, lee2022fusion, kosta2022adaptive} include preliminary evaluation of ANN and SNN inference energy based on \cite{horowitz20141}. This evaluation provides an incomplete analysis as it compares the computation energy but does not account for the cost of data movement, a significant contributor to inference energy in modern machine learning workloads \cite{kwon2019understanding}. Therefore, for our analysis, we use an analytical performance architecture, Timeloop \cite{parashar2019timeloop}, which incorporates both communication and computation energies. We adopt the same approach as in the study by \cite{sharma2022identifying} and map our networks to a specialized ML accelerator, Eyeriss \cite{chen2016eyeriss} to obtain the energy estimates. For spiking layers, as part of computation energy, only accumulator energy is considered as opposed to multiply-and-accumulate energy for ANNs. We used 45nm technology node for all our energy calculations.

\subsection{DSEC-flow Results}
\label{dsec}

\textbf{Ablation Study}: We perform an ablation study to verify our hypothesis (Section~\ref{intuition}) on the number and position of spiking layers in the hybrid architecture. We start with FireFlowNet architecture from Fig.~\ref{architecture}(b) and evaluate it on the DSEC-flow dataset. As depicted in Fig.~\ref{ablation}(a), we observe that the AEE for the FireFlowNet architecture increases as the number of spiking layers increases. We also conducted experiments with two spiking layers, exploring various layer positions (Fig.~\ref{ablation} (c)). The AEE for all the cases with two spiking layers is higher compared to the AEE with just one spiking layer. Moreover, in both single and two spiking layer cases, shifting these layers deeper into the network resulted in an increased AEE (Fig.~\ref{ablation} (b) and (c)).

All these ablations are following our hypothesis from Section~\ref{intuition}. Based on these findings, we conclude that keeping the first layer as a spiking layer in the hybrid architecture gives the lowest AEE due to reduced training complexity.  For all the forthcoming experiments, the first layer in the hybrid architecture is assigned to be the spiking layer, and the remaining layers consist of neurons with ReLU activation.

\begin{table}[htb]
    \caption{AEE (lower is better) on DSEC-flow. Best in bold.}
    \label{table1}
    \centering
    
    \begin{tabular}{c c c}
        \hline 
        \textbf{Network Architecture} &  \textbf{AEE} &  \textbf{Energy ($mJ$)} \\
        \hline\hline
        \cite{ponghiran2022event} (LSTM) & 1.28 & -\\
        EV-FlowNet \cite{gehrig2021raft} & 2.32 & - \\
        E-RAFT \cite{gehrig2021raft} & \textbf{0.79} & -\\
        \hline
         Base-Full-ANN &  1.71 & 962.13\\
         Base-Full-SNN & 1.47 & 1228.60\\
         \rowcolor{LightCyan} Base-Hybrid (ours) & 0.88 & 961.39\\
         \hline
         Mini-Full-ANN &  1.77 & 338.93\\
         Mini-Full-SNN & 1.65 & 508.96\\
         \rowcolor{LightCyan} Mini-Hybrid (ours) & 0.97 & 338.34\\ 
         \hline
         Micro-Full-ANN &  1.93 & 80.39\\
         Micro-Full-SNN & 1.76 & 144.26\\
         \rowcolor{LightCyan} Micro-Hybrid (ours) & 1.10 & \textbf{79.96}\\ 
         \hline
         FireFlowNet-Full-ANN &  6.56 & 352.74\\
         FireFlowNet-Full-SNN & 3.76 & 460\\
         \rowcolor{LightCyan} FireFlowNet-Hybrid (ours) & 3.14 & 356.21\\ 
         \hline

    \end{tabular}
    \vskip -0.01 in

\end{table}

\begin{figure}[t]
\centering
\includegraphics[width=0.40\textwidth]{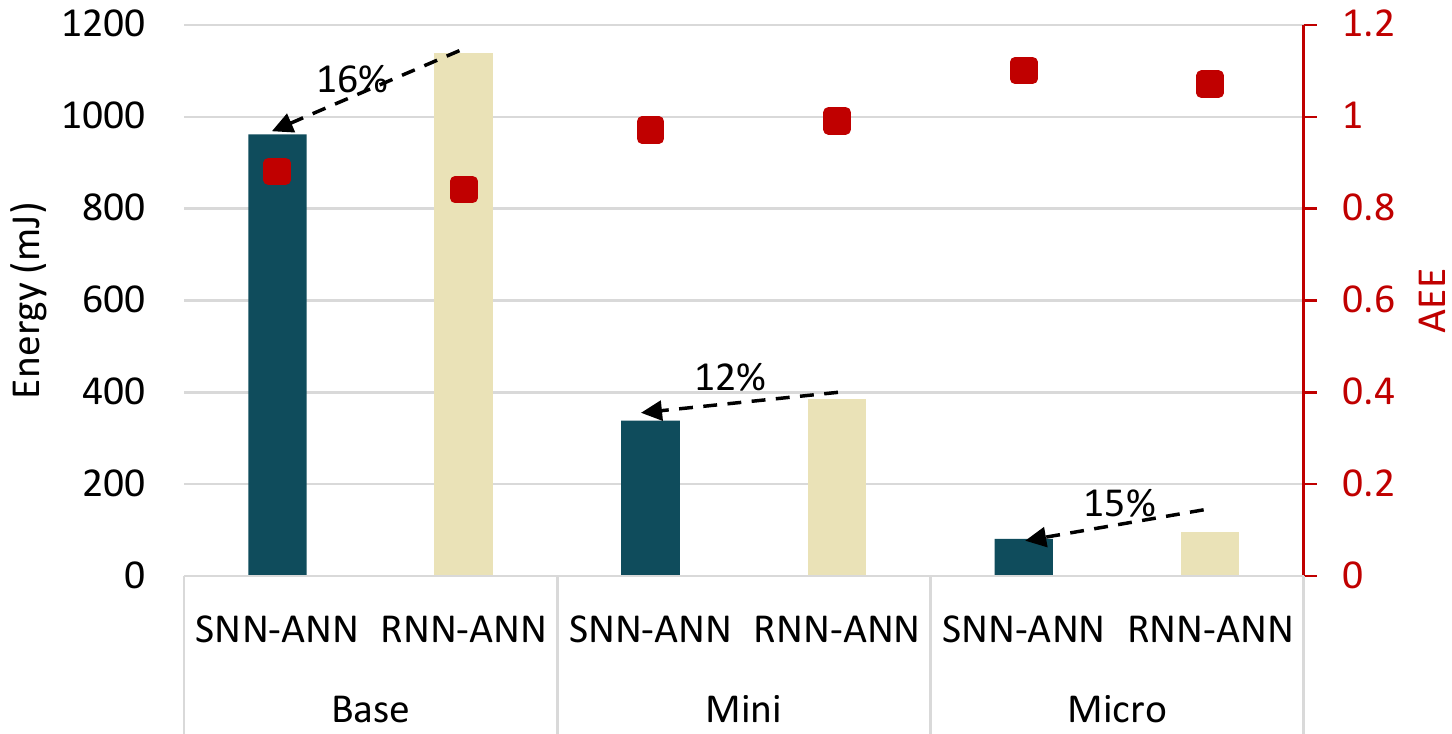}
\caption{AEE and Energy comparison to hybrid RNN-ANN\footref{fn2}.}
\centering
\label{comparison_rnn}
\vskip -0.2 in
\end{figure}

\noindent\textbf{Experiments with Network Size}: In this study, we explore the effect of reducing the number of base channels (k) in the architecture from Fig.~\ref{architecture}(a), and assess the performance of different network sizes using the training methodology from section~\ref{exp_setup}. We experiment with several values for k, such as Base (k=64), Mini (k=32) and Micro (k=16). Additionally, we compare our hybrid architecture with a Full-SNN (m=total number of layers), and Full-ANN (m=0), where m is the number of spiking layers.

\textbf{Full-SNN vs Full-ANN}: The results presented in Table~\ref{table1} show that all the Full-SNN architectures (Base - FireFlowNet) give lower AEE compared to the Full-ANN version. Specifically, Base-Full-SNN has ~14\% lower AEE compared to Base-Full-ANN. This improvement can be attributed to the membrane potentials in SNNs that enable them to learn the temporal information from the input. In contrast, ANNs being stateless, struggle to capture this temporal aspect since it is not encoded directly in the input. However, SNNs consume more inference energy compared to ANNs due to the overhead of fetching and updating the membrane potentials.

\textbf{Hybrid vs Full-SNN, Full-ANN}: Furthermore, we observe that the hybrid architecture outperforms the Full-ANN and Full-SNN architectures in terms of AEE. In addition to the low AEE, it also gives comparable inference energy to the Full-ANN architecture. Specifically, Base-Hybrid has 48\% lower AEE compared to Base-Full-ANN and $1.3\times$ lower energy than Base-Full-SNN. The lower AEE can be attributed to the fact that the spiking layer in the hybrid architecture can learn the temporal information from the input and also facilitate the optimizer to converge faster due to fewer trainable parameters. Additionally, the hybrid architecture consumes less energy than the Full-SNN architecture, because it has fewer spiking layers and therefore less overhead from membrane potentials. Another interesting observation is comparing architectures of different sizes. For instance, Micro-Hybrid has 35\% lower AEE and $12\times$ lower energy compared to Base-Full-ANN. Compared to Base-Full-SNN, Micro-Hybrid has 25\% lower AEE and $15\times$ lower energy.

\textbf{Hybrid vs state-of-the-art (SOTA)}: When compared to SOTA architectures for the DSEC flow dataset, our proposed Base-Hybrid architecture outperforms most of them. For example, in \cite{ponghiran2022event}, authors use LSTM layers instead of convolutional layers in the EV-FlowNet architecture, but our Base-Hybrid architecture shows a 31\% lower AEE than \cite{ponghiran2022event}. However, E-RAFT gives 12\% lower AEE compared to Base-Hybrid which can be attributed to the Gated Recurrent Unit and iterative flow refinement in E-RAFT. Note, however, that the complex architecture of E-RAFT would lead to higher inference energy compared to Hybrid architecture. 

\textbf{Hybrid SNN-ANN vs Hybrid RNN-ANN:} We also compare the hybrid SNN-ANN architecture with a corresponding hybrid RNN-ANN architecture which uses the ConvRNN layer (Fig.~\ref{architecture}(c)) as the first layer and all the other layers are ANN-like in EV-FlowNet architecture. We can see from Fig.~\ref{comparison_rnn} that as we reduce the model size, the AEE for the hybrid architecture is close to RNN with nearly 15\% lower energy. This can be attributed to the simpler inherent recurrence provided by SNNs compared to RNNs.

\begin{table*}[t]
    \vskip -0.05 in
    \caption{AEE and Energy results on MVSEC dataset. We report average AEE since there are multiple test sets. Best in bold. Fusion-FlowNet uses input from both the event and frame camera.}
    \label{table2}    
    \centering

    \begin{tabular}{c c c c c c c}
        \hline
        \textbf{Network Architecture} &  \textbf{outdoor\_day1} &  \textbf{indoor1}&  \textbf{indoor2}&  \textbf{indoor3}&  \textbf{Average AEE} &  \textbf{Energy ($mJ$)} \\
        \hline\hline
        EV-FlowNet \cite{evflow} &  0.49 &  1.03 &  1.72 &  1.53 &  1.19 & -\\
        $\text{Back to Event Basics}_{\text{EVF}}$ \cite{fireflownet} &  0.92 &  0.79 &  1.40 &  1.18 &  1.07 & -\\
        $\text{Back to Event Basics}_{\text{FIRE}}$ \cite{fireflownet} &  1.06 &  0.97 &  1.67 &  1.43 &  1.28 & -\\         
        XLIF-EV-FlowNet \cite{hagenaars2021self} &  0.45 &  0.73 &  1.45 &  1.17 &  0.95 & -\\
        XLIF-FireNet \cite{hagenaars2021self}  &  0.54 &  0.98 &  1.82 &  1.54 &  1.22 & -\\
        \hline
        Spike-FlowNet \cite{lee2020spike} &  0.49 &  0.84 &  1.28 &  1.11 &  0.93 & -\\
        Fusion-FlowNet \cite{lee2022fusion} &  0.59 &  \textbf{0.56} &  \textbf{0.95} &  \textbf{0.76} &  \textbf{0.72} & - \\       
        \hline
         Base-Full-ANN &  0.51 &  1.24 &  2.11 &  1.87 &  1.43 & 518.58\\
         Base-Full-SNN & 0.51 &  0.77 &  1.21 &  1.05 &  0.88 & 738.38\\
         \rowcolor{LightCyan} Base-Hybrid (ours) &  \textbf{0.4} &  0.69 &  1.05 &  0.92 &  0.77 & 518.21\\
         \hline
         Mini-Full-ANN &  0.56 &  1.23 &  2.13 &  1.90 &  1.45 & 207.63\\
         Mini-Full-SNN & 0.53 &  0.89 &  1.40 &  1.25 &  1.02 & 303.12\\
         \rowcolor{LightCyan} Mini-Hybrid (ours) & 0.41 &   0.69 &  1.04 &   0.92 &   0.77 & 207.27\\ 
         \hline
         Micro-Full-ANN &  0.74 &  1.24 &  2.14 &  1.90 &  1.50 & 49.04\\
         Micro-Full-SNN & 0.61 &  0.89 &  1.36 &  1.20 &  1.02 & 87.06\\
         \rowcolor{LightCyan} Micro-Hybrid (ours) & 0.42 &  0.79 &  1.20 &  1.07 &  0.87 & \textbf{48.78}\\       
         \hline
         FireFlowNet-Full-ANN &  1.15 &  1.35 &  2.20 &  1.96 &  1.66 & 194.39\\
         FireFlowNet-Full-SNN & 0.67 &  0.99 &  1.58 &  1.36 &  1.15 & 211.91\\
         \rowcolor{LightCyan} FireFlowNet-Hybrid (ours) & 0.59 &  0.91 &  1.50 &  1.27 &  1.07 & 196.29\\ 
         \hline

    \end{tabular}

    \vskip -0.2 in
\end{table*}

\subsection{MVSEC Results}
\label{mvsec}

We conducted experiments on the MVSEC dataset, using the same architectures employed for the DSEC-flow dataset. The quantitative results are presented in Table~\ref{table2} and visualized in Fig.~\ref{qualitative}. 

\noindent\textbf{Hybrid vs Full-SNN, Full-ANN}: The hybrid architecture outperforms the Full-ANN and Full-SNN architectures in terms of AEE and inference energy. Specifically, the  Mini-Hybrid showed 47\% and 25\% lower average AEE compared to the Full-ANN and Full-SNN versions, respectively. Furthermore, Mini-Hybrid demonstrated a 32\% reduction in inference energy compared to Mini-Full-SNN and had comparable energy consumption to Mini-Full-ANN. The ease of training the hybrid architecture contributes to the AEE improvement. Additionally, the reduction in inference energy compared to Full-SNN is due to the fewer spiking layers in the hybrid architectures

\noindent\textbf{Hybrid vs SOTA}: Mini-Hybrid, Mini-Full-SNN has a lower average AEE compared to the baselines that encode temporal information in the input \cite{fireflownet,evflow}. We attribute these improvements to the LIF neurons, which help to learn temporal information better compared to ANN architectures. In comparison to Spike-FlowNet \cite{lee2020spike}, which uses all the encoder layers as spiking layers, Mini-Hybrid has a 17\% lower average AEE. This finding is consistent with our ablation study, where we observed that the AEE tends to increase with an increase in the number of spiking layers (Fig.~\ref{ablation} (a)). Further, compared to Fusion-FlowNet \cite{lee2022fusion} Mini-Hybrid has only 6\% higher AEE. However, Fusion-FlowNet uses inputs from both the frame and event camera. In addition, we found that Base-Full-SNN and Mini-Hybrid achieve 7\% and 23\% lower average AEE, respectively than the fully spiking architecture XLIF-EV-FlowNet. We also compared the performance of the lightweight fully spiking architecture XLIF-FireNet \cite{hagenaars2021self} with our FireFlowNet-Full-SNN and Hybrid architecture and found that they achieve 6\% and 12\% lower average AEE, respectively. This performance of FireFlowNet-Full-SNN and FireFlowNet-Hybrid can be attributed to the presence of the BNTT \cite{bntt} and low training complexity in these architectures.

\section{Discussion}
\label{discussion}
Previous studies often reported lower accuracy for SNNs compared to ANNs \cite{rathi2020enabling, chowdhury2022towards, kim2023exploring}. These studies typically presented results on static datasets like CIFAR-10, CIFAR-100 \cite{krizhevsky2009learning} and ImageNet \cite{deng2009imagenet}. Optical flow estimation, which requires capturing temporal information from the input, benefits from SNNs' inherent recurrence (membrane potential). This enables SNNs to surpass ANNs in learning such temporal information while utilizing fewer parameters than RNNs. 
Furthermore, it is notable that inference energy is dominated by both computation and data movement \cite{kwon2019understanding}. Nonetheless, prior studies solely compared the computational energy of SNNs and ANNs. Therefore, for a fair comparison in this work, we consider the energy stemming from both computation and data movement. We observe that SNNs have nearly 30-40\% higher inference energy compared to ANNs when deployed on non-spiking hardware (Eyeriss). On the other hand, RNNs incur even more energy costs due to their considerably more complex computations. In light of these observations, hybrid SNN-ANN architecture offers an appealing middle ground (both for GPU implementation and training cost) by presenting itself as a low-energy alternative to Full-SNNs and RNNs while retaining their capabilities of processing temporal inputs effectively.

\section{Conclusion}
\label{conclusion}
In conclusion, this work presents a new approach to learning temporal information from event-inputs by introducing hybrid SNN-ANN architecture. Hybrid architectures benefit from the strengths of both SNN and ANN layers and can effectively learn the temporal information from inputs, without requiring any explicit input encoding. 
Our study demonstrates that a well-designed hybrid SNN-ANN architecture, with a spiking layer as the first layer and subsequent layers as ANNs, achieves better AEE compared to past hybrid architectures with multiple spiking layers. Additionally, this hybrid architecture alleviates the challenges of its deployment on non-spiking hardware (GPUs and custom ML hardware). 
Extensive quantitative and qualitative evaluations conducted on two commonly used optical flow datasets (DSEC-flow and MVSEC) demonstrate the superiority of hybrid architecture compared to state-of-the-art models. Finally, our deployment on non-spiking ML hardware shows that the inference energy of hybrid architecture is better than SNNs and RNNs and is comparable to ANNs.
We believe our work presents a promising direction for the development of efficient and accurate architectures with low training complexity and seamless deployment on traditional machine learning hardware.

\section{Acknowledgments}
This work was supported in part by the Center for Brain-inspired Computing Enabling Autonomous Intelligence (C-BRIC), one of six centers in JUMP, a Semiconductor Research Corporation (SRC) program sponsored by DARPA, in part by the National Science Foundation, in part by Intel, in part by IARPA MicroE4AI, in part by ARO W911NF-19-2-0237 and in part by the Vannevar Bush Faculty Fellowship.

{\small
\bibliographystyle{ieee_fullname}
\bibliography{main}
}

\end{document}